\documentclass[letterpaper, 10 pt, conference]{ieeeconf}  

\IEEEoverridecommandlockouts                              
\usepackage{graphics} 
\usepackage{wrapfig}
\usepackage{subcaption}
\usepackage{url}
\usepackage{epsfig} 
\usepackage{multicol}
\usepackage{color}

\usepackage{epsfig}
\title{\LARGE \bf
Development of a robotic system for \\automatic organic chemistry synthesis
}

\author{Joyce Xin-Yan Lim$^{1}$, Dasheng Leow$^{2}$, Quang-Cuong Pham$^{1}$, Choon-Hong Tan$^{2}$
\thanks{$^{1}$School of Mechanical and Aerospace Engineering, Nanyang Technological University, Singapore}
\thanks{$^{2}$School of Physical and Mathematical Sciences, Division of Chemical and Biological Chemistry, Nanyang Technological University, Singapore}
}

\begin{document}

\maketitle
\thispagestyle{empty}
\pagestyle{empty}

\begin{abstract}
  Automated chemical synthesis carries great promises of safety,
  efficiency and reproducibility for both research and industry
  laboratories. Current approaches are based on specifically-designed
  \emph{automation} systems, which present two major drawbacks: (i)
  existing apparatus must be modified to be integrated into the
  automation systems; (ii) such systems are not flexible and would
  require substantial re-design to handle new reactions or
  procedures. In this paper, we propose a system based on a
  \emph{robot arm} which, by mimicking the motions of human chemists,
  is able to perform complex chemical reactions without any
  modifications to the existing setup used by humans. The system is
  capable of precise liquid handling, mixing, filtering, and is
  flexible: new skills and procedures could be added with minimum
  effort. We show that the robot is able to perform a Michael
  reaction, reaching a yield of 34\%, which is comparable to that
  obtained by a junior chemist (undergraduate student in Chemistry).
\end{abstract}

\section{Introduction}

Automated chemical synthesis carries great promises of safety,
efficiency and reproducibility for both research and industry
laboratories~\cite{c1}. Recent contributions to this field range from
optimizing reactions using Artificial Intelligence (AI) \cite{c2}, 
full automation of drug syntheses \cite{c3}, discovery of new reactions or 
drugs using AI \cite{c4}, \cite{c5}, and flow chemistry in the industry \cite{c6}.

Most existing approaches are based on \emph{automation} systems, i.e.,
custom-built fixtures and mechanisms that perform a specific
experiment \cite{c3}, \cite{c4} and \cite{c6}. Such systems present two
main drawbacks. First, existing apparatus, such as mass spectrometers,
must be modified to be integrated into the automation systems. This
prevents those (usually expensive) pieces of equipment to be used by
other systems or by the chemists on other experiments. Second, such
systems are not flexible and would require substantial re-design to
handle new reactions or procedures. In some works, robot arms are
used, but they are confined to simple pick-and-place tasks \cite{c2},
\cite{c3}.

\begin{figure}[htp]
  \centering
  \includegraphics[width=0.49\textwidth, height=2in]{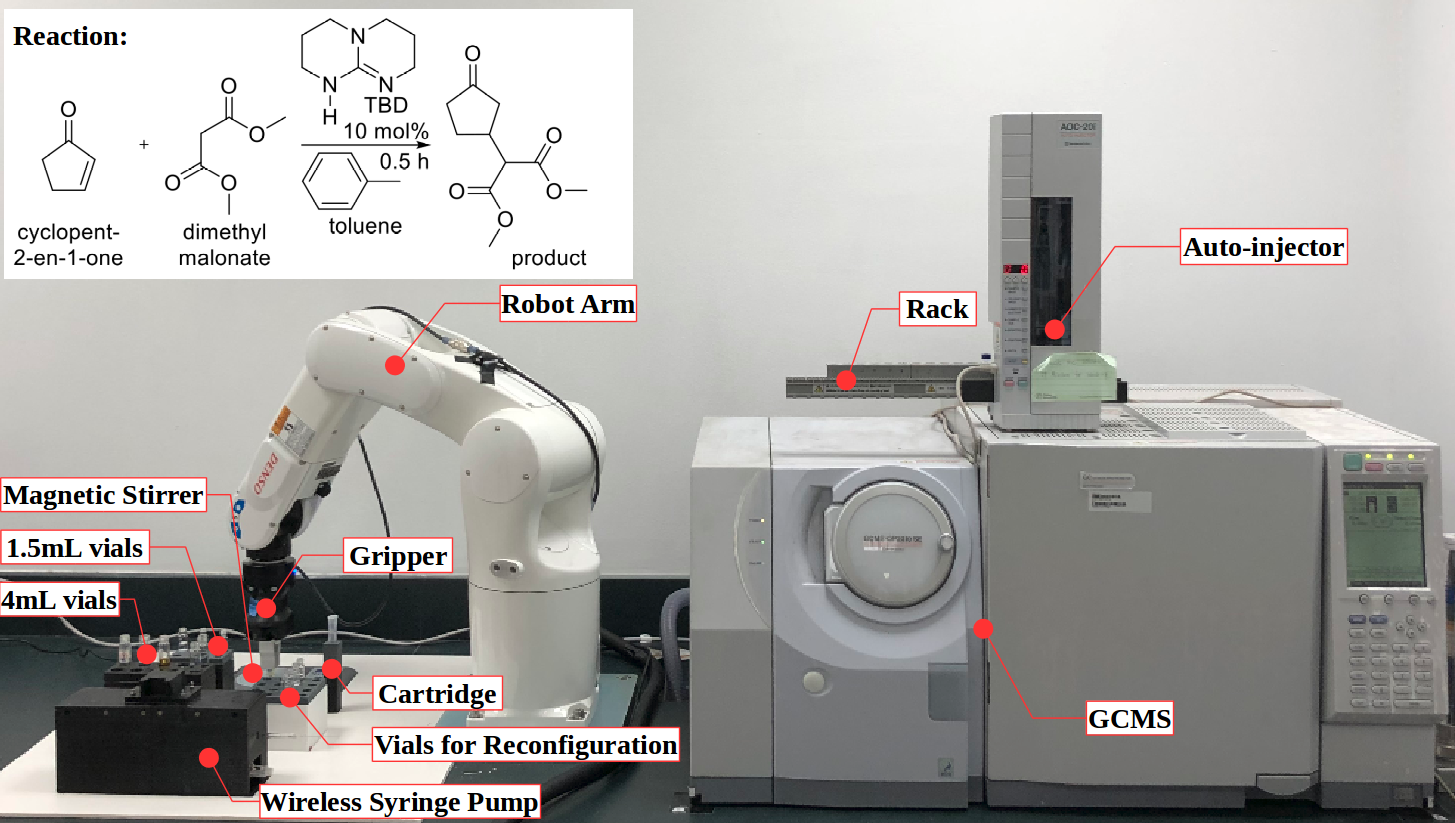}
  \caption{A robotic system capable of performing a complex,
    real-world organic chemistry reaction. The system -- consisting of
    a robot arm, a parallel gripper, a wireless electronic syringe
    pump -- is highly flexible and can work with existing equipment
    and setup without requiring any modifications. A video of the
    actual experiment can be viewed at
    \protect\url{https://youtu.be/xrPSiZuUhV8}}
  \label{fig:visual-id}
\end{figure}

In this paper, we propose a system based on a \emph{robot arm} which,
by mimicking the motions of human chemists, is able to perform complex
chemical reactions without any modifications to the existing setup
used by humans. In contrast with existing works, our robot arm
autonomously performs \emph{all} the steps of the experiment: liquid
handling, mixing, filtering, vial transfer, etc. This allows using
existing equipment and setup (vials, stirrer, mass spectrometer)
without any modifications, in the same spirit as our previous work on
the assembly of household furniture \cite{c7}.

We propose to demonstrate the robot capabilities in a real organic
chemistry experiment: the Micheal reaction. This reaction, a 1,4
addition to an $\alpha,\beta$-unsaturated carbonyl compound
(Fig.~\ref{fig:visual-id}), is an important reaction in organic
chemistry. We follow the same steps and use the same setup as human
chemist, more specifically:
\begin{enumerate}
\item \textit{Liquid handling}: The robot grabs the syringe pump and
  uses it to transfer liquids into a 1.5mL reaction vial;
\item \textit{Mixing of chemicals}: The robot picks the vial and
  brings it to a magnetic stirrer;
\item \textit{Filtering}: The robot pours the chemicals from
  a 1.5mL vial into a 3mL cartridge that is packed with silica and
  cotton for filtration. The cartridge is placed above an empty 1.5mL
  vial to store the filtered analyte sample;
\item \textit{Integrating with GCMS}: The robot brings the filtered
  analyte sample onto the rack of an auto-injector. which is mounted
  on a Gas Chromatography Mass Spectrometry (GCMS) machine;
\item \textit{Reconfiguration}: The robot reconfigures the environment
 in order to repeat the experiment;
\item \textit{Analysis}: The auto-injector administers analyte samples
  of the two runs into the GCMS for analysis to determine the yield.
\end{enumerate}

The remainder of the paper is organized as follows. In
Section~\ref{sec:related}, we review previous works on automated
chemical synthesis. In Section~\ref{sec:system}, we presents the
development of our system, which comprises: a robot arm, a parallel
gripper, a specially-designed syringe pump. The syringe pump is a
highlight of our setup: it was developed so as to deliver liquids with
very high precision, while being manipulated by the robot in a
flexible manner. We also show how to plan complex sequences of motions
for the robot to perform the chemical experiment in a very fast
manner. In Section~\ref{sec:experiment}, we describe the experiment
and report the results, including the comparisons with junior and
senior human chemists. Finally, in Section~\ref{sec:conclusion}, we
conclude and sketch some directions for future research.

\section{Related Work}
\label{sec:related}

Li et al. \cite{c2} created an autonomous chemistry laboratory coupled
with optimization algorithms to self-adjust the input parameters for
given targets to derive new nucleation theories. A robot arm was used
for object seeking and transferring of samples into the reaction
chamber or the workstation. For liquid injection, a peristaltic pump
that was installed on the reaction chamber was utilized.

Godfrey et al. \cite{c3} proposed an automated chemical synthesis
laboratory which is capable of modular expansion and allows chemists
to remotely guide chemical syntheses for drug discovery. Similar to
\cite{c2}, the robot arms were mostly used for pick-and-place
operations, but also include diluting of samples which was briefly
stated.

Williams et al. \cite{c8} developed a Robot Scientist to automate
early stages of drug design, by randomly screening a subset of its
library to filter suitable compounds and use AI to hypothesize
Quantitative Structure Activity Relationship (QSAR). Resembling
\cite{c2} and \cite{c3}, the system included robot arms and they were
used to transfer plates between different pieces of laboratory
automation equipment.

In these papers \cite{c2}, \cite{c3} and \cite{c8}, it showed that
robot arms were largely used to aid the operation of automation such
as pick-and-place tasks, but the chemical syntheses were only
conducted by the automation machines and not done by the robot arms.

The automation approach for these papers \cite{c4}, \cite{c5} and
\cite{c9} are quite similar as they made a single modular unit and
pumps, valves and tubes were used instead of robot arms. For
\cite{c4}, \cite{c5}, both studies had defined a chemical space or
connectivity model consisting of a pool of reagents which were linked
to the reactor while for \cite{c9}, the solutions were loaded into
cartridges that will be loaded to the synthesizer. The difference for
each study is that, Steiner et al. \cite{c4} built a general synthesis
unit that has a chemical programming language, which converts a
typical reaction scheme into an executable code, that run sequential
syntheses to produce drugs. Although Li et al. \cite{c9} also aim to
have a generalized synthesizer, they specifically target the synthesis
of small-molecules instead of drugs. On the other hand, Granda et
al. \cite{c5} used random combinations of starting materials to
perform reactions and then used AI to search for new reactivity within
the chemical space.

Adamo et al. \cite{c6} demonstrated an industrial continuous flow
manufacturing of Active Pharmaceutical Ingredient (API) that consisted
of reconfigurable modules. The unit was differentiated into two
streams, the upstream was for performing reactions and the downstream
was for purification and formulation of the API. Continuous flow is
created through pure automation and makes use of feeds and pumps to
transfer chemicals, which is also evident in \cite{c10}. The continuous
flow study conducted by Bedard et al. \cite{c10} was capable of
optimizing a specific reaction or a sequence of reactions, synthesize
a range of substrates under user-selected conditions or scale-up a
previous optimized synthesis.

To automate the experiments in these papers \cite{c4}, \cite{c5},
\cite{c8} and \cite{c9}, modifications such as sensors and tubing were
made to most apparatus to facilitate liquid transfer and all
components of the automation unit were dependent on each other. Hence,
it will not be convenient to dismantle the unit to access any of these
components individually due to the amount of tubing used. This
accessibility may be desired by research chemists who share a common
pool of equipment as by isolating equipment in the automation unit, it
can result in a smaller amount of resources available. In addition,
based on these studies, it is evident that most research involves pure
automation and there were not many implementations of robotics. Robot
arms were also mostly limited to pick-and-place tasks. However, robot
arms may pose several advantages due to their flexibility and may be
able to eliminate messy tubing to transfer liquids or
apparatus. Hence, robot arms may allow the integration of independent
equipment to automate chemical synthesis, such that the accessibility
of these individual equipment can be retained.

\section{System Development}
\label{sec:system}

\subsection{Overall System}

A Denso VS-060 robotic arm coupled with a Robotiq Hand-E gripper was
operated in a structured environment to conduct the experiment. An
automated and remote syringe pump was fabricated to perform liquid
handling and a GCMS with model, Shimadzu GCMS-QP2010SE, was used for
the analysis of the samples created by the robot.

\subsection{Development of Syringe Pump}

\begin{figure}[htp]
    \centering
	\includegraphics[width=0.45\textwidth, height=1.8in]{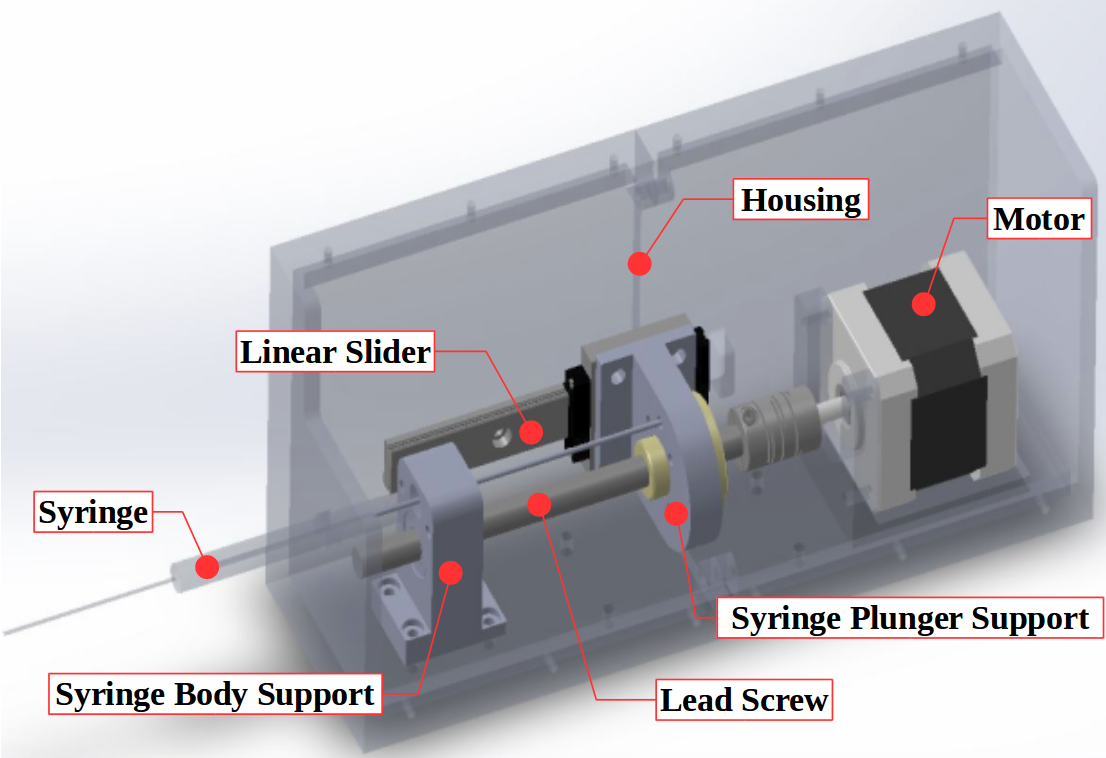}
    \caption{Syringe pump assembly}
    \label{fig:syringe}
\end{figure}

Our syringe pump was developed based on the foundations of previous
studies \cite{c12}, \cite{c13}. Both papers adopted a similar
approach, which was driving a stepper motor with a lead screw
assembly. The motor was driven by a motor driver however, an Arudino
micro-controller was used in \cite{c13} whereas a Raspberry Pi was
utilized in \cite{c12}. They were both wired to a computer for control
and both papers reported that their systems had high accuracy. In
\cite{c12}, it was reported that the accuracy of a NEMA11 motor was
+/-1\% and for the NEMA17 motor was +/-5\% measured in 1mL increments
and precision was relatively insensitive to microstepping for both
motors. In \cite{c13}, the study showed that a smaller syringe
capacities had better performance as they resulted in smaller flow
rate errors, and the flow rate errors for 0.5mL, 5mL and 60mL ranges
from 0.5\% to 3\%. The main difference of our syringe pump design 
as compared to the previous studies in \cite{c12}, \cite{c13} was that 
our design was remote whereas their designs had to be wired to a computer. 
In addition, our syringe pump was enclosed and contained the battery,
electronics and mechanical components. This design also included a
connection point so that the robot can grab the syringe pump and use
it (Fig.~\ref{fig:syringe}). An external computer controlled the
operations of both the robot and syringe pump via TCP communication
such that the motions of both systems can be synchronized.

We created a similar system as \cite{c12}, \cite{c13} and tested it 
on a 100$\mu$L syringe, using a NEMA17 motor with Arduino Uno and a 
Raspberry Pi 2. It was observed that the Arduino system took 1s whereas 
the Raspberry Pi system took 23s to draw 10$\mu$L when driven at 1/16
microstep. Thus, we decided to use a micro-controller instead of the
Raspberry Pi in our design due to better performance. We also tested
another micro-controller, the WeMos D1 Mini and noted that it achieved
the same result as the Arduino. However, the WeMos D1 Mini was much
smaller in size and had an ESP8266 module, thus enabling WiFi and
Transmission Control Protocol (TCP) communication. Hence, it fits our
system perfectly because the system had to be remote and battery
operated for the robot arm to use.

The pump consisted of 100$\mu$L glass syringe and a bipolar NEMA17
stepper motor with 0.9$^\circ$ step angle with a lead screw
assembly. The motor was controlled by a WeMos D1 Mini micro-controller
and a Big Easy motor driver. A DC power shield was added to the
micro-controller so that a 11.1V lithium ion battery could be used to
power both the micro-controller and the motor driver which were placed
in parallel. All connecting components were 3D printed and the
assembly is shown in Fig. 2.

The motor was controlled at 1/16 microstep to reduce vibration and
increase the resolution of the syringe. In \cite{c12}, the study
mentioned that accuracy of the assembly was found to be relatively
insensitive to microstepping. Hence, these indicated that the syringe
was able to accurately draw liquid at a better resolution than
chemists. This enabled the experiment to be scaled down which reduced
the amount of chemicals needed and a single syringe was able to cater
to the required capacity.

\subsection{Motion Planning Procedure}

The robot arm had numerous goal targets to reach to perform the
experiment and each target had multiple solutions for the joint
angles. As such, we require a sequence that could result in the
shortest total trajectory time for all targets. A two-step algorithm,
the Robotic Task Sequencing Program (RoboTSP), had been proposed by a
previous study \cite{c14}, to determine a near-optimal order of
configurations for the manipulator to visit for \textit{n} goal
points, with \textit{m} robot configurations per target. The outline
of RoboTSP was to first, determine the visit sequence in \textit{task
  space} and next, a graph search will be conducted to find the
shortest path in \textit{configuration space} based on the visit
sequence.

However, the chemical experiment was made up of several sequential
tasks and each task had \textit{n} goal targets. As these sequential
goal targets had a defined order, only the second step of RoboTSP was
extracted to obtain the optimal sequence in the configuration space.

\begin{figure}[htp]
    \centering
	\includegraphics[width=0.49\textwidth, height=1.9in]{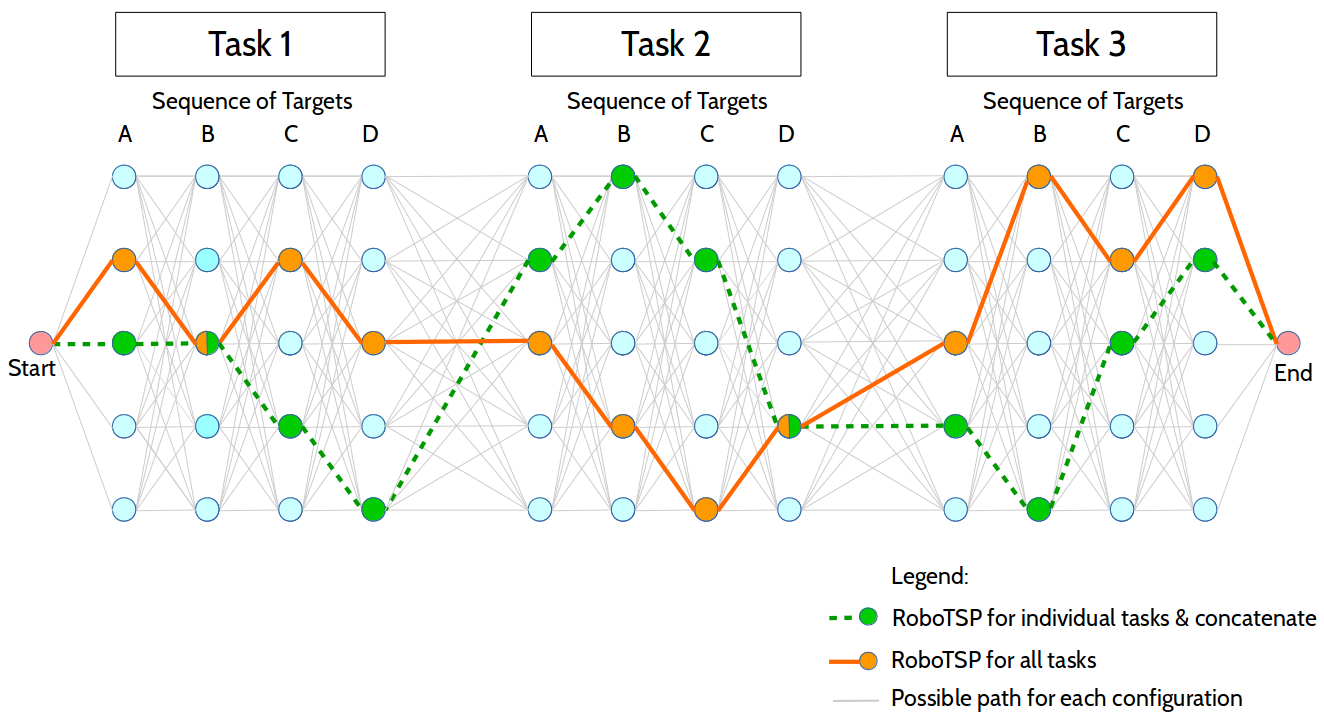}
        \caption{Obtaining optimized order of configurations. The
          circles depict \textit{m} robot configurations for
          \textit{n} targets of each task and although \textit{m=4}
          and \textit{n=5}, it was observed that the number of
          configurations were not constant for all targets during the
          actual planning.}
        \label{fig:rtsp}
\end{figure}

We came up with two different methods to acquire the optimal sequence
by using RoboTSP as explained below (see Fig.~\ref{fig:rtsp}):
\begin{enumerate}
\item Get the optimal sequences of each individual task and
  concatenate them based on order of tasks. This meant that for
  \textit{x} tasks, RoboTSP was ran \textit{x} times.
\item Concatenate the targets based on the order of tasks to obtain
  the optimal sequence in one sitting. In this case, RoboTSP was only
  ran once.
\end{enumerate}

It was observed that the final sequence in Method 1 (dotted line)
differs from Method 2 (continuous line), although there were some
common configurations within the both methods, as illustrated in
Fig. 3.

\subsection{System Integration}
A system with Intel i7 2.20GHz processor and 8GB RAM, running Ubuntu
16.04 (Xenial) was used in this experiment. ROS was used to control
both the Denso VS-060 robotic arm and the Robotiq Hand E gripper. The
motion planning and collision checking was done in OpenRAVE.

\section{Experiments}

The Michael reaction protocol that was used in the experiment was
extracted from \cite{c11}. The reaction was carried out in a 1.5mL
vial where toluene, cyclopent-2-en-1-one, dimethyl malonate were
reacted using TDB catalyst, and the reaction diagram can be seen in
Fig. 1. The chemicals were stirred and subsequently added to ethyl
acetate. The mixture was poured into a cartridge packed with silica
powder for filtration and more ethyl acetate was poured to flush out
the analyte sample.

\label{sec:experiment}

\begin{figure*}[htp] 
  \begin{subfigure}[b]{0.5\linewidth}
    \centering
    \includegraphics[width=0.95\linewidth,height=5cm]{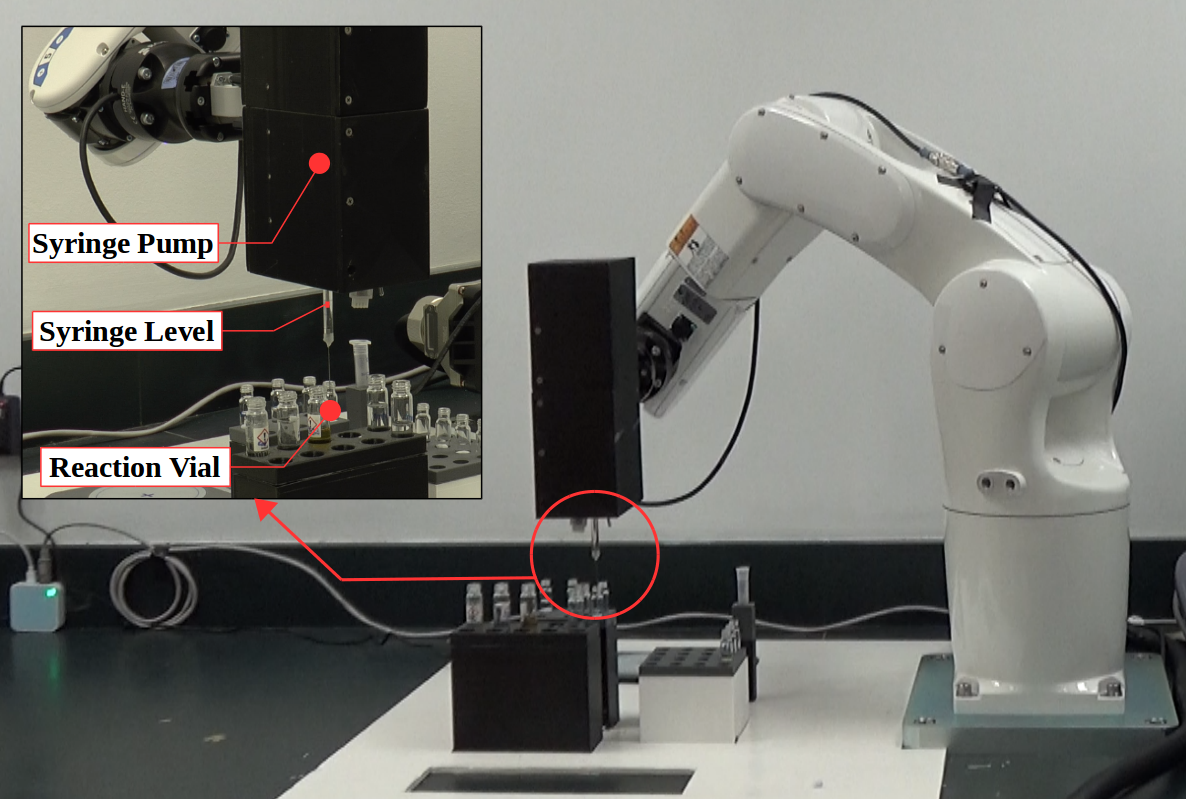} 
    \caption{Using syringe pump to handle liquids} 
    \vspace{4ex}
  \end{subfigure} 
  \begin{subfigure}[b]{0.5\linewidth}
    \centering
    \includegraphics[width=0.95\linewidth,height=5cm]{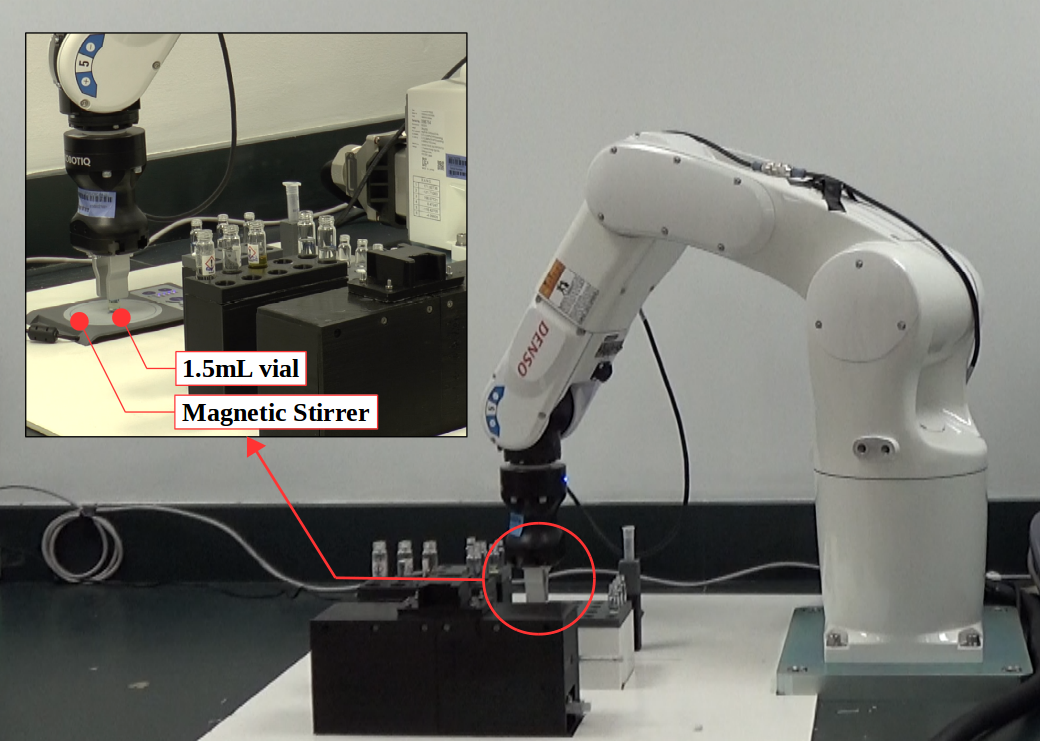} 
    \caption{Stirring of chemicals} 
    \vspace{4ex}
  \end{subfigure} 
  \begin{subfigure}[b]{0.5\linewidth}
    \centering
    \includegraphics[width=0.95\linewidth,height=5cm]{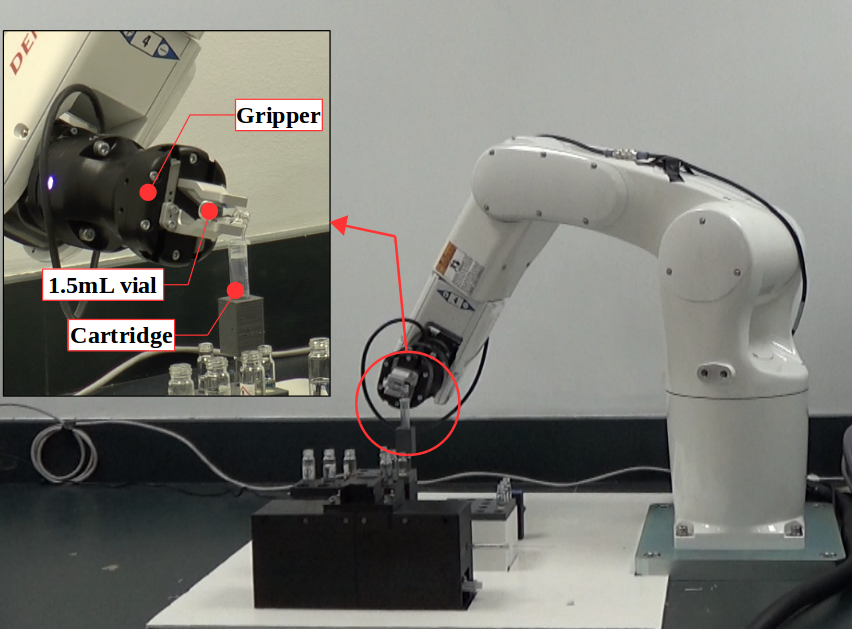} 
    \caption{Filtering chemicals by pouring} 
    \vspace{4ex}
  \end{subfigure}
  \begin{subfigure}[b]{0.5\linewidth}
    \centering
    \includegraphics[width=0.95\linewidth,height=5cm]{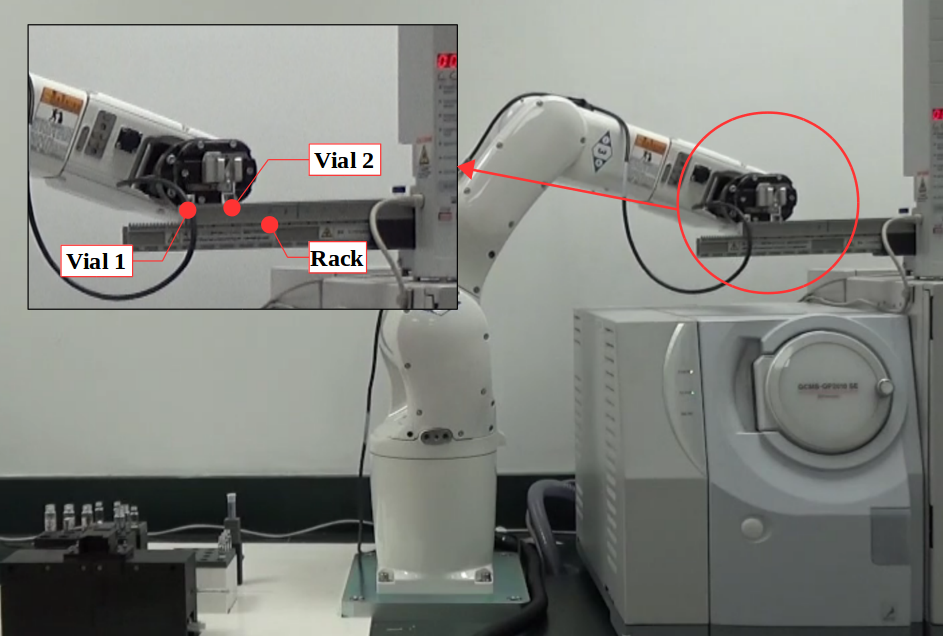} 
    \caption{Placing vial on auto-injector rack} 
    \vspace{4ex}
  \end{subfigure} 
  \caption{Snapshots of the robot carrying out the experiment. The
    full video of the experiment can be viewed at
    \protect\url{https://youtu.be/xrPSiZuUhV8}}
  \label{fig:snapshot}
\end{figure*}

\subsection{Procedure}
As presented in Section I, the detailed steps of the experiment are
listed below (see Fig.~\ref{fig:snapshot}):

\paragraph{Liquid Handling} The arm grabbed the syringe pump and used
it to move reactants into a 1.5mL reaction vial, as shown in Fig
4a. Each time after a chemical was transferred, the syringe was
flushed with acetone 5 times to wash it before a different chemical
can be drawn by the syringe.

\paragraph{Mixing of chemicals} The arm picked the reaction vial,
which contains a magnetic stir bar, and brought it to a magnetic
stirrer to mix the chemicals, as shown in Fig 4b. The arm held the
vial securely above the stirrer for 5 minutes and mixing was required
to carry out the reaction. Once the reaction was completed, the arm
brought the vial back onto the rack. Due to the dexterity of the
robot, it eliminated the need of using a clamp to secure the vial
above the magnetic stirrer.

\paragraph{Transferring reaction mixture} The arm grabbed the syringe
pump again and used it to add reaction mixture from the 1.5mL reaction
vial to another 1.5mL vial containing ethyl acetate. This was to
dilute the analyte sample to meet the concentration requirements of
the GCMS and also to standardize the volume of reaction mixture used
for analysis in each run.

\paragraph{Mixing of chemicals} The vial with ethyl acetate and
reaction mixture also contains a magnetic stir bar, and the arm
brought the vial to the magnetic stirrer for 5 minutes.

\paragraph{Filtering} Filtration was required to remove the catalyst
from the mixture. The arm grabbed a silica-packed 3mL cartridge which
was seated in a holder, and placed it on top of an empty 1.5mL
vial. Filtration was done by using the arm to pour chemicals from the
vial containing reaction mixture into the 3mL cartridge, as shown in
Fig 4c. The arm then waited 5 minutes for the filtration by gravity to
be completed. An additional 1.5mL vial with only ethyl acetate was
poured into the cartridge as well to flush out the analyte sample. The
arm waited another 5 minutes for the filtration to be completed. The
trajectory for pouring was done by bringing the end-effector to the
rim of the cartridge, and applying Cartesian twists to the
end-effector to create the pouring action. The twists were applied
gradually, which meant that the pouring action did not only make up of
a single twist, but multiple twists. As one of the vials used for
pouring contains a magnetic stir bar while the other vial did not, the
Cartesian twists applied were different because the stir bar was
obstructing the flow. Thus, the twists had to be more gradual for the
vial with the stir bar to prevent spillage even though the total
twists for both vials were rather similar. The total twist applied for
the first vial with the stir bar was at 122$^\circ$ and
\textit{$\Delta$x=203mm} towards the cartridge whereas for the second
vial, the angle was 118$^\circ$ and \textit{$\Delta$x=201mm}. These
twists were determined through experimenting.

\paragraph{Integrating with GCMS} The arm brought the filtered analyte
sample onto the rack of an auto-injector which was mounted on the
GCMS, as shown in Fig 4d. The trajectory this step was done by
planning with constraints on the \textit{x, y, z} axes of the gripper
as the vial was not capped. Hence, these restrictions on the
end-effector was to prevent obtaining a trajectory that could cause
the vial to turn upside down and result in spillage.

\paragraph{Reconfiguration} The arm reconfigured the environment and
Steps 1-5 were repeated. The newly filtered analyte sample was placed
on another hole of the rack of the auto-injector.

\paragraph{Analysis} The auto-injector was activated and it
administered both analyte samples into the GCMS for analysis to
determine the yield of the sample. The GCMS took 30 minutes including
cooling, to analyze each run.

Hence, each experiment consisted of 2 runs, which yielded 2 vials of
analyte samples. The only difference between Run \#1 and Run \#2 was
the extra reconfiguration step which was present in Run \#2.

\subsection{Specifications}

The specifications for each run are listed in Table I. The offline
planning time was the total time taken to obtain the optimized
sequence from RoboTSP, which was discussed in Section III-C, and plan
trajectories to these configurations for all tasks.

The number of planned trajectories corresponds to the number of joint
configurations from the optimized sequence, however the actual number
of trajectories executed were more than the number of planned
trajectories, as some trajectories were repeated during washing of
syringe in the liquid handling step. The actual execution time
consisted of the time taken for the robot to execute trajectories and
the time taken for the syringe to draw and push liquid, which did not
include waiting time. As the total waiting time consisting of the
stirring time and filtration time amounted to 20 minutes, the total
time for Run \#1 and Run \#2 was approximately 30 minutes and 31.5
minutes respectively.

\begin{table}[h]
\caption{Experimental specifications for each run}
\begin{center}
\begin{tabular}{|l|r|r|}
\hline
 Specifications & Run \#1 & Run \#2 \\
\hline
Offline Planning Time (s)& 17	& 20 \\
\hline
No. of Tasks & 19	& 20 \\
\hline
No. of Planned Trajectories & 109 & 151 \\
\hline
Actual Execution Time (minutes)& 10	& 11.5 \\
\hline
\end{tabular}
\end{center}
\end{table}

\subsection{Results}

\paragraph{Motion Planning Results} As discussed in Section III-C,
there were two proposed methods in obtaining the optimal sequence. We
did a comparison on planning duration for 5 tasks between both
methods, as shown in Table II.

\begin{table}[h]
\caption{Comparison on planning duration for 5 tasks}
\begin{center}
\begin{tabular}{|c|c|c|}
\hline
  & Planning Time (s) & Trajectory Duration (s) \\
\hline
Method 1 & 4.1 & 88.4 \\
\hline
Method 2 & 2.3 & 70.0 \\
\hline
\end{tabular}
\end{center}
\end{table}

The planning time is the total time taken to plan to all joint
configurations of every task and the trajectory duration is the total
time taken to execute all trajectories of every tasks. The planning
time and trajectory duration in Method 1 was longer by 77\% and 26\%
respectively, as compared to in Method 2. This indicated that the
sequence obtained from Method 1 was less optimal as compared to Method
2. Therefore, we used Method 2 to obtain the optimal sequence and
compute the offline planning time in Section IV-B.

\paragraph{Synthesis Results} The GCMS was ran under the following
conditions:
\begin{itemize}
    \item Restek Rtx-5MS, 30m x 0.23 mmID, x 0.25 $\mu$m df column.
    \item Hold at 60$^\circ$C for 3 minutes and heat at 6$^\circ$C/min
to 180$^\circ$C.
    \item Pressure at 56.7kPa, column flow 0.99mL/min, split injection
mode, split ratio:100.
    \item Injection port temperature at 250$^\circ$C.
    \item Ion source temperature at 200$^\circ$C, pressure at 56.7kPa,
m/z 60-250.
\end{itemize}

\begin{table}[h]
\caption{Analysis of samples from GCMS}
\begin{center}
\begin{tabular}{|l|r|r|}
\hline
  & Concentration (mg/mL) & Yield (\%) \\
\hline
Experiment 1, Run \#1 & 6.47	& 24 \\
\hline
Experiment 1, Run \#2 & 8.11	& 30 \\
\hline
Experiment 2, Run \#1 & 9.90	& 37 \\
\hline
Experiment 2, Run \#2 & 8.30	& 31 \\
\hline
Junior chemist & 9.89	& 36 \\
\hline
Senior chemist & 14.45	& 54 \\
\hline
\end{tabular}
\end{center}
\end{table}

The analyte samples created by the robot in Experiments \#1 and \#2
were sent for analysis. From the analysis in Table III, it proved that
the robot was capable of conducting chemical synthesis. The average
yield for Run \#1 of both experiments was 30.5\%, and the average
yield for Run \#2 was also 30.5\%. In addition, the overall mean yield
across both experiments was 30.5\% too. The concentration is directly
proportional to the yield and is dependent only on the yield. Hence,
the average concentration across all runs was 8.2mg/mL of analyte
sample.

We also showed that the robot arm had the ability to perform
reconfiguration and repeat the experiment, which was due to its
dexterity. This indicates the possibility of implementing AI into the
experiment to optimize reactions. Once an optimized reaction had been
found, the robot would be able to reproduce a similar optimized result
by repeated the experiment, due to its ability to achieve consistent
results.

We also compared the yield obtained by the robot to those obtained by
a junior chemist (undegraduate student in Chemistry) and a senior
chemist (postdoctoral researcher). The chemicals used were of the same
batch of chemicals that was used in Experiment 2 conducted by the
robot.

\subsection{Discussion}

From the analysis in Table III, it showed that the robot was capable
of achieving consistent yield for both runs of each experiment, even
though the Experiments 1 and 2 were conducted on separate instances
and a new batch of chemicals was used for each experiment. The batch
of chemicals that were used evidently affects the yield to a certain
extent which can be seen by comparing Experiments 1 and 2. The runs in
Experiment 2 had a generally higher yield as compared to Experiment 1
although the average yield across the run numbers were the same.

The analysis also illustrated that the robot could obtain a similar
yield as a junior chemist (34\% versus 36\%). The comparison was done
in this matter such that it would be fair, since the batch of
chemicals used can affect the results to a certain extent, which was
discussed in the previous analysis. However, we observed that the
yield achieved by the senior chemist was higher than that of the
robot, which may be due to the high skills of the senior chemist,
achieved through years of training. One possible improvement could be
for example in the liquid transfer step: during the transfer,
chemicals may stick onto the walls of the vials, leading to lesser
amount of reactants, hence a lower yield. Understanding the skills of
the senior chemist and transferring them to the robot would be a very
exciting direction of research.

\section{Conclusions}
\label{sec:conclusion}

We have looked into the feasibility of automating chemical synthesis
using a robot arm and achieved an average yield of 30.5\% and an
average concentration of 8.2mg/mL of analyte sample formed. The experiment
proved that robot arms have the ability in automating experiments and are
not limited to pick-and-place tasks which was seen in previous studies. Taking 
advantage of the dexterity of the robot, we were able to integrate independent 
systems together to carry tasks to complete the experiment. The tasks executed 
by the robot include using the syringe pump to transfer liquids, doing pouring, 
bringing the reaction vial to the stirrer and to the auto-injector rack. In 
conclusion, based on the advantages attributed to robot arms such as consistency 
of results and abilities to integrate independent systems together, robot arms can 
be used to automate chemical syntheses. 

We envision three main lines of future research. First, we can improve the yield 
and repeatability of the reactions by better understanding the skills of the senior 
chemist and transferring them to the robot. Second, we can add new skills to the robot, 
in order to perform more complex reactions, by implementing mechanisms for learning
from human demonstration. Finally, as we have demonstrated that the robot can autonomously
perform two runs, it can be easily extended to perform multiple runs. This opens the
possibility of using Machine Learning techniques to learn optimal reaction parameters 
by performing the experiment multiple times. Our future work will explore this exciting avenue.



\section*{Acknowledgment}
This work is supported by Nanyang Technological University (M4012040) and Ministry of Education.


\end{document}